\documentclass[unnumsec,webpdf,contemporary,large]{oup-authoring-template}%

\graphicspath{{Fig/}}

\theoremstyle{thmstyleone}%
\theoremstyle{thmstyletwo}%
\theoremstyle{thmstylethree}%

\begin{document}

\journaltitle{Journal Title Here}
\DOI{DOI HERE}
\copyrightyear{2022}
\pubyear{2019}
\access{Advance Access Publication Date: Day Month Year}
\appnotes{Paper}

\firstpage{1}


\title{Prot2Chat: Protein LLM with Early-Fusion of Text, Sequence and Structure}

\author[1]{Zhicong Wang}
\author[2]{Zicheng Ma}
\author[1,$\ast$]{Ziqiang Cao}
\author[1]{Changlong Zhou}
\author[2]{Jun Zhang}
\author[2]{Yiqin Gao}


\address[1]{\orgdiv{School of Computer Science and Technology}, \orgname{Soochow University}, \orgaddress{\postcode{215006}, \state{Suzhou}, \country{China}}}
\address[2]{\orgdiv{Changping Laboratory}, \orgname{Peking University}, \orgaddress{\postcode{102206}, \state{Beijing}, \country{China}}}

\corresp[$\ast$]{Corresponding author. \href{email:email-id.com}{zqcao@suda.edu.cn}}

\received{Date}{0}{Year}
\revised{Date}{0}{Year}
\accepted{Date}{0}{Year}



\abstract{
\textbf{Motivation:} Proteins are of great significance in living organisms. However, understanding their functions encounters numerous challenges, such as insufficient integration of multimodal information, a large number of training parameters, limited flexibility of classification-based methods, and the lack of systematic evaluation metrics for protein Q\&A systems. To tackle these issues, we propose the Prot2Chat framework.\\
\textbf{Results:} We modified ProteinMPNN to encode protein sequence and structural information in a unified way. We used a large language model (LLM) to encode questions into vectors and developed a protein-text adapter to compress protein information into virtual tokens based on these vectors, achieving the early fusion of text and protein information. Finally, the same LLM reads the virtual tokens and the questions to generate answers. To optimize training efficiency, we froze the encoder and employed Low-Rank Adaptation (LoRA) techniques for the LLM. Experiments on two datasets show that both automated metrics and expert evaluations demonstrate the superior performance of our model, and zero-shot prediction results highlight its generalization ability. The models and codes are available at \url{https://github.com/wangzc1233/Prot2Chat}. \\
\textbf{Contact:} \href{name@email.com}{zqcao@suda.edu.cn} or \href{name@email.com}{wangzc025@163.com}
}



\maketitle

\section{Introduction}
With the continued advancement of biotechnology, gaining a deeper understanding of protein sequences, structures, and their functions has become increasingly critical. 
Protein function is fundamentally governed by its amino acid sequence, which encodes the structural and biochemical features necessary for activity. However, even identical protein sequences can adopt distinct three-dimensional conformations under varying environmental conditions or subtle mutations, thereby exhibiting diverse biological functions, as exemplified by cases reported in AFcluster~\cite{wayment2024predicting}. The widespread availability of protein sequence and structure data—enabled by high-throughput sequencing technologies and powerful structure prediction tools~\cite{abramson2024accurate} has significantly advanced our understanding of the physicochemical properties of proteins. In parallel, significant research efforts have focused on investigating and predicting protein functions, where deep learning models have employed sequence or structure representations as input to predict function-related properties such as Gene Ontology (GO) terms~\cite{kulmanov2020deepgoplus, su2023saprot}.
The recent emergence of large language models (LLMs) has propelled protein question answering (Q\&A) systems at the forefront of computational biology research~\cite{zhou2023survey}.
The related systems enable researchers to efficiently access and interpret key information regarding protein properties and functions. This holds great potential for transformative applications in fields such as drug discovery and disease research~\cite{smith2025funcfetch}.
To support this growing field, several datasets, such as Mol-Instructions~\cite{fang2023mol} and UniProtQA~\cite{luo2023biomedgpt}, have been introduced, providing essential resources for advancing protein Q\&A systems.
An increasing number of models have fine-tuned LLMs on protein-specific Q\&A datasets to generate contextually relevant text or to answer questions related to protein structure and function~\cite{zhou2025decoding}.

However, as summarized in Table~\ref{tab:base models}, existing methods face several critical limitations, including inadequate integration of multimodal information, restricted flexibility, high training parameter volume, and a lack of systematic evaluation metrics.
These challenges constrain the models’ ability to generate accurate and context-aware responses.
For instance, although models such as FAPM~\cite{xiang2024fapm} and InstructProtein~\cite{wang2023instructprotein} incorporate generative Q\&A capabilities, they remain heavily dependent on classification-based paradigms.
This often leads to the production of unexpected content, deviating from the intended scope of inquiry and reducing the models' effectiveness in handling complex protein-related questions. BiomedGPT~\cite{luo2023biomedgpt} enables users to interact with biological data through natural language. Nonetheless, it relies exclusively on sequence-level information, which substantially limits its predictive capacity. ProtChatGPT~\cite{wang2024protchatgpt} and Evola~\cite{zhou2025decoding} attempt to address this limitation by employing separate modules for encoding sequence and structural information.
However, this modular design hinders effective interaction between sequence and structural features and significantly increases computational complexity. Moreover, no existing model has systematically examined the relative contributions of sequence and structural information to the performance of protein Q\&A systems.

\begin{figure}[t]
  \includegraphics[width=0.98\linewidth]{sample.jpg} 
  \caption {Prot2Chat can assist human in understanding protein information and achieve cross-modal information communication. Among them, `\(<\)Soft Prompt\(>\)` is the prompt obtained by our model through fusing protein structure, sequence, and text information, which helps the LLM generate more valuable answers.}
  \label{fig:sample}
\end{figure}

As the sequence and structural features of proteins are in inseparable correspondence, both jointly determine the properties of proteins. 
We propose a novel framework named Prot2Chat. 
This framework can seamlessly integrate the spatial structure and sequence information of proteins into LLMs. 
Meanwhile, by incorporating text information during the processing of protein information, the LLM is enabled to understand the question, so as to achieve question-based compression of protein information, and ultimately realize the early fusion of text, structure, and sequence.
This approach seeks to enhance both the performance and applicability of protein Q\&A systems. 
To be specific, we introduce protein sequence information during the initialization of nodes features in ProteinMPNN~\cite{dauparas2022robust} and early-fuse the protein sequence information in the protein feature representation.
Then inspired by BLIP-2~\cite{li2023blip}, an adapter module was incorporated to align the protein encoder with a LoRA~\cite{hu2022lora} fine-tuned LLM. 
In addition, we use the LLM to encode the question text and add it to the adapter, integrating it into the above-mentioned compression process to further achieve the early-fusion of protein and text information, aiming to improve the accuracy and robustness of protein Q\&A.
The processed protein and text data are provided as a soft prompt to the LLM. 
Leveraging the flexible text generation capabilities of the LLM, our model effectively follows residue-level protein instructions, significantly enhancing its utility in protein-related tasks. 
As shown in Figure \ref{fig:sample}, the answering effect of our model can provide effective insights for human to explore proteins in a Q\&A manner familiar to them.

To evaluate the model's performance, we conducted comparative experiments on the Mol-Instructions~\cite{fang2023mol} and UniProtQA~\cite{luo2023biomedgpt} datasets, and performed zero-shot experiments on a subset of UniProtQA to validate generalization. 
We used common metrics like BLEU and ROUGE, as well as online KIMI~\cite{qin2024mooncake} scoring metrics and expert evaluations, ensuring the reliability and relevance of the answers. 
Experimental results show that Prot2Chat which integrates protein sequences and structures using a unified encoder and fuses text information at an early stage outperforms those using sequence-only pretraining, ESM-based encoders~\cite{lin2022language}, and multi-encoders. 
Besides, its effectiveness far outperforms that of late-fusion, which means that text information is only utilized when the LLM generates answers.
These findings highlight the effectiveness of early-fusion strategies for integrating multimodal protein representations.
Furthermore, the high consistency between expert evaluations and online KIMI assessments confirms the robustness of Prot2Chat.

In summary, our contributions are as follows:

\begin{enumerate}[1.]
\item We extended the existing structure encoder ProteinMPNN to realize structure and sequence early-fusion without the need for training.
\item Based on this protein encoder, we designed a text-aware protein-text adapter module. 
By re-utilizing the LLM, we achieved question-based compression of protein information, early-fusion of text and protein information, and implemented a lightweight and efficient protein LLM with only 109 million training parameters.
\item We performed various systematic assessments across various evaluation datasets to validate our model's generative and generalization capabilities. These included traditional metrics, large model evaluations online, and manual expert evaluations.
\end{enumerate}

\section{Related work}

\subsection{Protein Representation Learning}
Proteins are fundamental components of cells, essential for their biological activities and diverse functions. Previous studies on protein characterization have explored various methods to learn protein representations based on different forms of protein information. Protein sequences, often referred to as the "language of life," have been extensively studied using advanced natural language processing techniques. For instance, Tranception~\cite{notin2022tranception} employs the Transformer~\cite{vaswani2017attention} model to encode amino acid sequences, capturing relationships between residues and autoregressively reconstructing protein sequences from large-scale databases. Similarly, sequential modeling approaches such as ESM~\cite{lin2023evolutionary} leverage masked language modeling (MLM) to develop attention patterns that correspond to residue-residue contact maps, enhancing sequence-based protein representations.
On the other hand, structure-based approaches~\cite{gligorijevic2021structure} directly indicate protein function and encode geometric information of the protein for topology-sensitive tasks such as protein property prediction. Foldseek~\cite{van2022foldseek} introduces the concept of a structural alphabet, encoding protein structures into a discrete representation space. Similarly, Saprot~\cite{su2023saprot} introduces a structure-aware vocabulary, embedding structural information into model input to enhance representational capacity, achieving significant success in protein function prediction tasks.

\subsection{Multimodal Alignment}

Enabling LLMs to understand extra modalities has been a fast-evolving research area, with notable examples including image-text models~\cite{li2023blip}, video-text models like VideoLlama~\cite{zhang2023video}, audio-text models such as Macaw-LLM~\cite{lyu2023macaw}, and molecular-text models like MolTC~\cite{fang2024moltc}. 
This line of research was pioneered by advancements in visual language modeling (VLM), to enable LLMs to understand images, leading VLM methods adopt different strategies. 
Some, like BLIP-2~\cite{li2023blip}, employ nonlinear and expressive cross-modal projectors, while others, like PaLM-E~\cite{driess2023palm}, utilize visual encoders and fine-tune LLMs on multimodal datasets. These approaches have also been increasingly applied to protein analysis. For instance, Galactica~\cite{taylor2022galactica} leverages scientific literature to model protein sequences and SMILES representations, enabling the model to interpret sequence properties. 
Some of these methods attempt to adopt a Q\&A approach that is closer to human habits to explore and understand the functions of proteins. 
ProtNote~\cite{char2024protnote}, utilize two encoders to encode protein sequence information and text information, respectively, and Multilayer Perceptron (MLP) to fuse these inputs, achieving the goal of using text to achieve supervised and zero-shot protein functional prediction. 
ProtChatGPT~\cite{wang2024protchatgpt} uses two independent modules to encode sequence and structural information respectively, and aligns them with natural language to explore protein functions. 
InstructProtein~\cite{wang2023instructprotein} employs a supervised training framework based on instruction generation using knowledge graphs, enabling bidirectional generation between natural language and protein language. This model can predict protein functional descriptions and generate protein sequences based on natural language prompts. Prot2Text~\cite{abdine2024prot2text} combines graph neural networks (GNNs) with LLMs to predict protein functions in free-text form. FAPM~\cite{xiang2024fapm} utilizes a contrastive learning framework to implement a generative-like Q\&A model, while BiomedGPT~\cite{luo2023biomedgpt} bridges the gap between biological and human natural language through a multimodal generative pre-trained transformer (GPT). This allows users to "communicate" with biological modalities using free-text queries, facilitating interactions with biological data in natural language. 

\begin{figure*}[t]
  \includegraphics[width=0.98\linewidth]{0418.jpg} 
  \caption {Model Structure of Prot2Chat. The red font represents the input, the snowflake represents freezing, and the flame represents the parameters to be trained. We obtain the embedding with multi-dimensional feature fusion of proteins from protein structure and sequence information through an encoder. Meanwhile, we get the question vector from the question text. Then, we conduct early-fusion and alignment of this vector with protein information to obtain the soft prompt. Finally, we input the soft prompt along with the question text into the LLM to get the answer.}
  \label{fig:model}
\end{figure*}

\begin{table*}[ht]
    \centering
    \caption{Comparison of input-output of protein Q\&A models, "Description" is a text related to protein functions, such as the description of Gene Ontology (GO) terms, "[Structure+Sequence+Question]" represents the early-fusion of structure, sequence and text. }
    \renewcommand{\arraystretch}{1.25}
    \begin{tabular}{l|ccr}
        \hline
        Model & Input & Output \\
        \hline
        Prot2Text & Sequence & Structured Text  \\
        ProtNote & Sequence\&Description & Function\&Probability    \\
        ProtChatGPT & Structure\&Sequence\&Question & Free Text  \\
        Evola & Structure\&Sequence\&Question & Free Text  \\
        InstructProtein & Sequence\&Question & Structured Text  \\
        FAPM & Sequence & Structured Text  \\
        BioMedGPT & Sequence\&Question & Free Text  \\
        Prot2Chat & [Structure+Sequence+Question]\&Question & Free Text  \\
        \hline
    \end{tabular}
    
    \label{tab:base models}
\end{table*}

\begin{table}[ht]
    \centering
    \caption{The count and division of the datasets we used.}
    \renewcommand{\arraystretch}{1.25}
    \begin{tabular}{l|cccr }
        \hline
        Dataset & Train & Valid & Test\\
        \hline
        Mol-Instructions & 404640 & 16859 & 11072 \\
        UniProtQA & 25820 & 1075 & 6734  \\
        \hline
    \end{tabular}
    
    \label{tab:dataset} 
\end{table}

    

\begin{table*}[!ht]
    \centering
    \caption{Result in Mol-instructions protein oriented dataset, the best performances are marked in bold.}
    \renewcommand{\arraystretch}{1.25}
    \begin{tabular}{l|c|c|c|c}
    \hline
        Model & BLEU-2 & ROUGE-1 & ROUGE-2 & ROUGE-L    \\ 
        \hline
        LLaMA3-8B-Instruct & 4.84 & 23.07 & 4.31 & 15.36  \\ \hline
        LLaMA3-FT with sequence& 6.42 & 24.50  & 6.32 & 17.03\\ \hline
        BioMedGPT-LM-10B & 1.02 & 10.93 & 1.57 & 7.84  \\ \hline
        Evola-10B & 8.69 & 29.09 & 8.41 & 20.04  \\ \hline
        KIMI (Zero-shot)& 4.79 & 22.21 & 4.62 & 14.70  \\ \hline
        KIMI (Few-shot) & 12.05 & 31.21 & 11.38 & 24.18   \\ \hline
        
        Prot2Chat  & \textbf{35.85}  & \textbf{57.21} & \textbf{38.09} & \textbf{50.51}  \\ \hline

    \end{tabular}
    
    \label{tab:result1}
\end{table*}

\begin{table*}[!ht]
    \centering
    \caption{Results of the ablation experiments in Mol-instructions dataset. The best performance is marked in bold.}
    \renewcommand{\arraystretch}{1.25}
    \begin{tabular}{l|c|c|c|c}
    \hline
        Model & BLEU-2 & ROUGE-1 & ROUGE-2 & ROUGE-L    \\ 
        \hline

        Prot2Chat  & \textbf{35.85}  & \textbf{57.21} & \textbf{38.09} & \textbf{50.51}  \\ \hline

        w/o Fine-tuned LLM& 12.87 & 30.89 & 14.67 & 26.81  \\ \hline
        w/o Protein sequence & 31.88 & 52.90 & 33.43 & 46.17 \\ \hline
        w/o Early-fusing text  & 33.25  &54.88& 35.26 & 47.90  \\ \hline
    \end{tabular}
    
    \label{tab:ablation}
\end{table*}

\begin{table*}[!ht]
    \centering
    \caption{Result in UniProtQA dataset, the best performances are marked in bold.}
    \renewcommand{\arraystretch}{1.25}
    \begin{tabular}{l|c|c|c|c}
    \hline
        Model & BLEU-2 & ROUGE-1 & ROUGE-2 & ROUGE-L    \\ 
        \hline
        LLaMA3-8B-Instruct & 2.41  & 13.66 & 1.10  & 9.07  \\ \hline
        BioMedGPT-LM-10B & 5.80  & \textbf{15.88} & 7.32 & 14.68  \\ \hline
        Prot2Chat (Zero-shot)& 2.23 & 14.73 & 1.73 & 10.10  \\ \hline
        
        Prot2Chat (Fine-tuned) & \textbf{6.72} & 15.71 & \textbf{9.25}  & \textbf{15.57}  \\ \hline
    \end{tabular}
    
    \label{tab:result2}
\end{table*}

\begin{table}[!ht]
    \centering
    \caption{Results of KIMI evaluation, "Avg." represents the average ranking.}
    \renewcommand{\arraystretch}{1.25}
    \begin{tabular}{l|l|l|l|l|c}
    \hline
        Model & 1st & 2nd & 3rd & 4th & Avg.  \\ \hline
        Prot2Chat & 386 & 242 & 13 & 9 & 1.45  \\ \hline
        BioMedGPT-LM-10B & 26 & 57 & 326 & 241 & 3.20  \\ \hline
        LLaMA3-FT with sequence  & 18 & 22 & 252 & 358 & 3.46  \\ \hline
        Evola-10B  & 220 & 329 & 59 & 42 & 1.88  \\ \hline
    \end{tabular}
    
    \label{tab:select1}
\end{table}

\begin{table}[!ht]
    \centering
    \caption{Results of expert manual evaluation, "Avg." represents the average ranking.}
    \renewcommand{\arraystretch}{1.25}
    \begin{tabular}{l|l|l|l|l|c}
    \hline
        Model & 1st & 2nd & 3rd & 4th & Avg.  \\ \hline
        Prot2Chat & 359 & 266 & 17 & 8 & 1.49  \\ \hline
        BioMedGPT-LM-10B & 28 & 58 & 334 & 230 & 3.18  \\ \hline
        LLaMA3-FT with sequence  & 11 & 21 & 239 & 379 & 3.52  \\ \hline
        Evola-10B  & 252 & 305 & 60 & 33 & 1.81  \\ \hline
    \end{tabular}
    
    \label{tab:select2}
\end{table}

\begin{table*}[t]
    \centering
    \caption{Case study. The bold font represents the key points of the answers, the blue font represents the correct answers of the key points, and the red font represents the incorrect answers of the key points.}
    \renewcommand{\arraystretch}{2} %
    \begin{tabular}{m{0.2\textwidth}|m{0.37\textwidth}|m{0.14\textwidth}|m{0.14\textwidth}}
        \hline
        Inputs & LLaMA3-FT with sequence & Prot2Chat & Ground Truth \\
        \hline
        \begin{minipage}[c][0.01\textwidth][c]{0.2\textwidth} %
            \centering
            \includegraphics[width=0.8\linewidth]{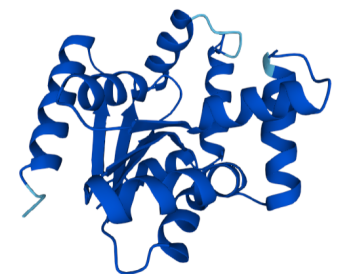} \\
            \textbf{[O58736]}Using the protein structure and the following protein sequence as input, assess its potential function by comparing it to known proteins with similar sequences or structures: 
        \end{minipage}  &The provided protein sequence is approximately 240 amino acids long and shows no significant similarity to known proteins in the database. However, it can be analyzed for potential functional regions and motifs using various bioinformatic tools. Here are some observations and predictions: 1. \color{red}Transmembrane regions\color{black}: The sequence contains several regions that are \color{red}predicted to be transmembrane,\color{black} including the following: 24-45 amino acids (ILLLVGMPGSGKGEVAKAFRRRGIPVI) 63-82 amino acids (MGGDAIREEAEKRGIP KTPEGLKYVSLKVREE) & With the provided structure and amino acid sequence, the protein is expected to \color{blue}display adenylate kinase activity and participate in the phosphorylation, ATP biosynthetic process, phosphorylation. & The protein with the structure and amino acid sequence has been analyzed, and its structure suggests a potential \textbf{involvement in ATP binding}.\\
        \hline
    \end{tabular}
    
    \label{tab:example}
\end{table*}

\section{Method}
\subsection{Overview of Prot2Chat}
As illustrated in Figure~\ref{fig:model}, the overall architecture of our model comprises three components: the protein encoder, the text-aware protein-text adapter, and the LLM. 
Our model leverages the 3D coordinate structure information of protein residue atoms along with the protein sequence and natural language question text. 
The protein encoder processes protein information to generate a protein embedding representation, which is then transformed into a text-aligned protein prompt via the text-aware protein-text adapter, with question text information being fused simultaneously. This prompt, combined with the input question, serves as the input to the large language model, ultimately producing the output answer.
\subsection{Sequence and Structure Fused Protein Encoder}
We modified the structure encoder ProteinMPNN to realize the sequence and structure early-fusion.
ProteinMPNN is originally dedicated to design protein sequences based on backbone structures. 
The input is structural information $E$, namely the 3D coordinates of protein residue atoms ($N$: amino nitrogen, $C\alpha$: alpha carbon, $C$: carbonyl carbon and $O$: carbonyl oxygen). 
Then protein features $h_V$ are obtained through the encoder layer and the decoder layer. 
\begin{align}
    h_E &= \text{Linear}(E) \\
    h_V &=\text{Zeros}(n,Dc) \label{EQ:originalHv}\\
    h_V, h_E &= \text{Encoderlayers}(h_V, h_E) \\
    h_V &= \text{Decoderlayers}(h_V, h_E) \label{EQ:origndecoderlayers}
\end{align}
where Linear, Encoderlayers and Decoderlayers are all modules in ProteinMPNN, $E$ is the structual features and $n$ is the number of protein residues and $Dc$ is the embedding dimension of the protein encoder.
We find that a variant version of ProteinMPNN additionally contains the Embedding module for protein sequences $S$, and the Decoderlayer is sequence-aware, which changes Equation~\ref{EQ:origndecoderlayers} to:
\begin{equation}
    h_V = \text{Decoderlayers}(h_V, h_E, \text{Embedding}(S))
\end{equation}
To be specific, the Decoderlayer combines the node embeddings, the output features of the encoder, and the sequence embeddings to construct the context features for decoding through neighborhood feature aggregation. 
Using the message passing mechanism, the hidden state $h_V$ is continuously updated through multilayer stacking to simulate the dynamic generation process of the decoder. 

On this basis, we further initialized $h_V$ with the off-the-shelf sequence embedding to make the sequence and structural information early-fused.
Specifically, Equation~\ref{EQ:originalHv} is replaced by:
\begin{equation}
    h_V =\text{Embedding}(S)
\end{equation}
In particular, all of the model weights we used are from ProteinMPNN, with no need for additional training. 
The final protein node vector ${h}_{V}$ is used as the protein feature embedding for the text-aware protein-text adapter. 
Inspired by InstructPLM~\cite{qiu2024instructplm}, we concatenate the protein representations encoded by nine released ProteinMPNN models. 
As a result, $Dc = 128*9 = 1152$.

\subsection{Text-Aware Protein-Text Adapter}

We implemented a text-aware protein-text adapter to semantically align the information obtained from the protein encoder with natural language, incorporating question text information simultaneously. Subsequently, this fused information is provided as a soft prompt to the LLM.
The protein embedding $h_V$ obtained from the modified ProteinMPNN model is output through the linear projection layer, positional encoding and the cross-attention mechanism. 
During the above-mentioned process, text information is fused simultaneously, as shown in Figure ~\ref{fig:adapter}.
\begin{figure}
    \centering
    \includegraphics[width=0.98\linewidth]{508.jpg}
    \caption{Detail of the Text-Aware Protein-Text Adapter. The adapter takes protein embedding and question vector as inputs. It integrates text information at an early stage through a set of learnable queries. Subsequently, using these queries, the protein embedding is used as keys/values for cross-attention calculations. This cross-attention module compresses the protein information into a fixed length and captures the key protein information according to the queries infused with question text information, resulting in soft prompt to assist the LLM in generating more accurate responses. }
    \label{fig:adapter}
\end{figure}

To be specific, the input protein feature first passes through a linear projection layer \( W_{\text{proj}} \), the aim is to transform the input features to the target output dimension $D_{\text{o}}$:
\begin{equation}
X_{\text{proj}} = W_{\text{proj}} \cdot h_V + PE
\end{equation}
where $PE$ is the Dynamic Positional Encoding~\cite{vaswani2017attention} used to capture the positional information of amino acids.

Meanwhile, we utilize the question text and the LLM to obtain the hidden state \(Qh_t\) of the last text token as the question vector, so as to achieve prompt compression and carry out the early-fusion of text and protein information, enabling the model to perceive question text information earlier: 

\begin{equation}
Qh_{\text{t}} = \text{LLM}(\text{question}).hiddenstate
\end{equation}

The complete protein encoding is too long. 
We adopt the idea of BLIP-2 to extract the important semantic features from it. 
In particular, we introduce \(n_\text{q}\) learnable queries \(\boldsymbol{Q} \in \mathbb{R}^{n_\text{q}\times D_\text{o}}\), incorporate the question vector \(Qh_t\) obtained from the question text, and also apply the positional encoding (PE) to them.

Then the multi-head cross-attention layer is adapted to capture key protein information based on queries: 

\begin{equation}
A^{k}=\text{softmax}\left(\frac{Q^{k}K^{k\top}}{\sqrt{D_{k}}}\right)V^{k}
\end{equation}
where $Q^{k}=\boldsymbol{Q}W_{Q}^{k}$, $ K^{k}=X_\text{proj}W_{K}^{k}$ and $V^{k}=X_\text{proj}W_{V}^{k}$ stand for queries, keys and values. 
Among them, $W_{Q}^{k},W_{K}^{k},W_{V}^{k}$ are related parameters, $k$ is the head index.

Then, the outputs of multi-head cross-attention are concatenated across heads and linearly transformed by ${W}_{out}$ into the final protein prompt integrated with text information, which serves as the soft prompt for the LLM, where $M$ is the number of attention heads:
\begin{equation}
\boldsymbol{X_\text{protein}}=\text{Concat}(\boldsymbol{A}^{1},\boldsymbol{A}^{2},\cdots,\boldsymbol{A}^{M})\boldsymbol{W}_{out}
\end{equation}

\subsection{LLM Decoder}

We combine $X_\text{protein}$, which integrates text and protein information, with the text question, and then input this combination into the existing LLM to obtain a response.
\begin{equation}
\text{response} = \text{LLM}(X_\text{protein}, \text{question})
\end{equation}
To improve domain adaptability, we fine-tuned the LLM with LoRA while training the adapter. 
The number of adapter training parameters is 106,483,712, while the number of LLaMA3 training parameters is 3,407,872. 

The total number of training parameters of the Prot2Chat is 109M. 
Meanwhile, the number of training parameters of BiomedGPT-10B is 3B, that of Evola-10B is approximately 1.7B, the range of trainable parameters of Prot2Text is from 256M to 898M, and that of FAPM is 188M. 
This also shows the advantage of our model in significantly reducing the computational cost. 
Besides, we adpot the CrossEntropyLoss function as common.

\section{Experiment}

\subsection{Datasets}
We evaluated the performance of our model using the Mol-Instructions~\cite{fang2023mol} and UniProtQA~\cite{luo2023biomedgpt} datasets. For our experiments, we utilized the protein-oriented instruction subset in Mol-Instructions, which is primarily derived from entries in the UniProtKB/Swiss-Prot database~\cite{uniprot2018uniprot}. 
This subset encompasses tasks such as predicting protein domains, functions, and activities. Our model was predominantly trained on this dataset. 

Additionally, we employed a portion of the UniProtQA dataset, introduced by BiomedGPT~\cite{luo2023biomedgpt}, to assess the generalization capability of our model through zero-shot evaluation. Fine-tuning was also performed on this dataset to further validate performance. UniProtQA consists of textual descriptions of proteins, including their functions and properties. 
The detailed sizes of our dataset splits are provided in Table~\ref{tab:dataset}.
It is worth noting that there are significant differences in the text style and length between the UniProtQA and Mol-Instructions datasets. This directly leads to large differences in the traditional metrics of the subsequent evaluation test results.
To incorporate structural and sequence information, we retrieved the corresponding PDB files for proteins listed in Swiss-Prot. These PDB files, combined with the associated questions, serve as the input to Prot2Chat.

\subsection{Baselines}

We introduced the following baselines that represent protein information using sequence.
\begin{itemize}
    \item \textbf{LLaMA3: } LLaMA3~\cite{dubey2024llama} series has attracted a lot of research attention due to its outstanding capabilities in the general domain. We use LLaMA3-8B-Instruct and input the protein sequence and question to achieve zero-shot.  

    \item \textbf{LLaMA3-FT: } We also fine-tuned LLaMA3-8B-Instruct using textual protein sequence information and used this model as a reference model.

    \item \textbf{BioMedGPT-LM-10B: } BioMedGPT~\cite{luo2023biomedgpt} is a domain-specific LLM trained on a large selection of corpora of human scientific knowledge. The model encodes the protein sequence with the ESM-3B~\cite{lin2022language} and uses BioMedGPT-LM-7B as the decoder to generate the response. 
    
    \item \textbf{Evola-10B: }  Evola~\cite{zhou2025decoding} is trained on hundreds of millions of protein Q\&A pairs and an AI-generated dataset of 150 billion words. It adopts Retrieval-Augmented Generation (RAG) to integrate external knowledge and performs excellently in generating accurate, detailed, and context-relevant answers about protein functions, providing expert-level insights for proteomics and functional genomics research. Here, we use its model with 10B parameters as the comparative model.
    
    \item \textbf{KIMI: } We used the online large model KIMI as a control.
\end{itemize}
In order to make a fair comparison, we used the same input template for each model through prompt engineering, so that the model could output the answer as we instructed. 
Some other open-source models in related fields, such as FAPM, Prot2Text, and InstructProtein, were not included in the comparison because these models complete protein Q\&A tasks based on classification, or there are differences in the formats of the model's inputs and outputs, as shown in Table \ref{tab:base models}. 
Therefore, it is hard to conduct comparisons. 
In addition, the ProtChatGPT model was not included in the comparison because accessible resources for this model could not be found.

\subsection{Model Setting}

We jointly trained the text-aware protein-text adapter and LLaMA3 on the Mol-Instructions dataset, employing full training for the adapter and LoRA fine-tuning for LLaMA3. 
Inspired by InstructPLM~\cite{qiu2024instructplm}, we configured the adapter with 256 queries and integrate text information on each query. 
For the LoRA setup, we set the rank \( r \) to 8, LoRA alpha to 16, and targeted the "q\_proj" and "v\_proj" modules, with a LoRA dropout rate of 0.1. The adapter comprises 106,483,712 trainable parameters, while LLaMA3 has 3,407,872 trainable parameters. 
We utilized the Adam optimizer and implemented gradient accumulation to optimize the training process. The initial learning rate was set to \(10^{-4}\), with a batch size of 2 and a maximum context length of 1024 tokens, which includes both the question and answer text. The model was fine-tuned for 2 epochs, with each training session requiring approximately 1600 hours on an NVIDIA RTX 3090 GPU.
\subsection{Evaluation Metrics}

To evaluate the effectiveness of the model's text generation, we employed performance metrics including BLEU~\cite{papineni-etal-2002-bleu} and ROUGE~\cite{lin-2004-rouge}. Additionally, the target text and the texts generated by different models are simultaneously input into the online KIMI model, using the following prompts: 
\textbf{Please select the sentence that is closest and most accurate in meaning to the Target sentence and rank sentences A, B, C and D accordingly, with the first place meaning the most identical and accurate in meaning to the Target. Only provide the answer, for example, `B D A C' or `A C D B', etc. 
Target: \{Ground Truth\} A: \{Model1 Generated Text\} B: \{Model2 Generated Text\} C: \{Model3 Generated Text\} D: \{Model4 Generated Text\}}
This allowed us to determine which generated answer was closer to the target text, thereby assisting in the evaluation of the model's output quality. Furthermore, we conducted expert manual evaluations, where professional biology PhDs ranked the responses of different models based on their alignment with the target text.

\subsection{Main Results}

The results of the assessment are shown in Table \ref{tab:result1} to Table \ref{tab:select2}. 
The BLEU-2 and ROUGE-L scores of Prot2Chat exceed those of models such as BiomedGPT and Evola. 
The results show that there is a significant modal gap between the protein sequence and the natural language, and it is incomprehensible to directly input the protein sequence as text and problem into the LLM, resulting in the disordered and meaningless response of the model. 
Aligning protein language with human language is an effective way to solve this problem. 
Moreover, we have observed that the performance of the model trained through the early-fusion of structural information and sequences is significantly superior to that of other models leveraging either sequences or structures.
This suggests that we can directly use the structure and sequence information of proteins to help language models better understand and generate technical terms through feature space alignment. 
In addition, the early-fusion of text information and protein information also demonstrates a significant effect.
Unification of multimodal protein information and establishment of a connection between proteins and natural language can provide guidance for researchers in their study of unknown proteins.
The results of our model's prediction on the UniProtQA~\cite{luo2023biomedgpt} dataset verify the generalization performance of our model. 
On the other hand, the evaluation results of online large language models, as shown in Table~\ref{tab:select1}, demonstrate that our model outperforms other comparative models overall and the results of expert manual evaluations (Table~\ref{tab:select2}) are consistent with those obtained using KIMI.
Further illustrating the role of our model in protein Q\&A tasks.

\subsection{Ablation Experiments}
As shown in Table~\ref{tab:ablation}, we carried out a comparative analysis to evaluate the impact of incorporating protein structure information during the training process. 
Comparing with the results of the model trained with only the sequence further highlights the importance of structural information for understanding proteins.
We also studied the effect of fine-tuning LLMs. The results show that training only the adapter is insufficient to achieve optimal performance.
By comparing the results obtained from training in the cases of early-fusion and non-early-fusion of text and protein information, we found that the early-fusion of information plays a non-negligible role in the understanding of proteins and the generation of large language models, and its effect is significantly better than that of late-fusion.
These findings emphasize the necessity of integrating protein structure information and a unified multimodal alignment module, which enables large language models to better understand and generate domain-specific technical terms.

\subsection{Case Study}
As shown in Table \ref{tab:example}, the results obtained from our use of the protein Q\&A test dataset indicate that the responses generated by the model trained by adding protein structure and early-fusing text information are significantly superior to those of other trained models (taking LLaMA3-FT as an example here). 
This further demonstrates that protein structure is of crucial importance for the comprehensive understanding of proteins, the early-fusion and alignment of multimodal information plays a non-negligible role in the understanding of proteins and the generation of large language models.

\section{Conclusion}
This paper introduces a novel protein Q\&A model that addresses the limitations of traditional protein Q\&A tasks by leveraging the modified ProteinMPNN and a text-aware adapter module to integrate protein structure and sequence information with natural language text. By harnessing the flexible generation capabilities of large language models (LLMs), our approach effectively bridges the gap between protein data and textual understanding. Experimental results demonstrate that the model trained by adding protein structure and early-fusing text information significantly outperforms other baseline models. This not only highlights the crucial role of protein structure in protein understanding and analysis but also proves the effectiveness of the early-fusion of protein information and text.
Furthermore, the results validate the effectiveness and generalization ability of our model in protein Q\&A tasks.

\bibliographystyle{plain}
\bibliography{reference}

\end{document}